%% file: main.tex
\input{_0_vars}

\ifdefined\isarxiv
\documentclass[11pt]{article}
\usepackage[numbers]{natbib}
\else
\documentclass{article} 
\usepackage{iclr2026_conference,times}

\fi

\ifdefined\isarxiv

\usepackage{amsmath}
\usepackage{amsthm}
\usepackage{amssymb}
\usepackage{algorithm}
\usepackage{subfig}
\usepackage{algpseudocode}
\usepackage{graphicx}
\usepackage{grffile}
\usepackage{wrapfig,epsfig}
\usepackage{url}
\usepackage{xcolor}
\usepackage{epstopdf}

\usepackage{bbm}
\usepackage{dsfont}

\else 

\usepackage{hyperref}
\usepackage{url}

\fi

\ifdefined\isarxiv

\else

\usepackage{amsmath}
\usepackage{amsthm}
\usepackage{amssymb}
\usepackage{algorithm}
\usepackage{subfig}
\usepackage{algpseudocode}
\usepackage{graphicx}
\usepackage{grffile}
\usepackage{wrapfig,epsfig}
\usepackage{url}
\usepackage{xcolor}
\usepackage{epstopdf}

\usepackage{bbm}
\usepackage{dsfont}

\fi
 
\allowdisplaybreaks

\ifdefined\isarxiv

\usepackage{tikz}
\usepackage{hyperref}  
\hypersetup{colorlinks=true,citecolor=blue,linkcolor=blue} 
\usetikzlibrary{arrows}
\usepackage[margin=1in]{geometry}

\else
\fi
 
\graphicspath{{./figs/}}

\theoremstyle{plain}
\newtheorem{theorem}{Theorem}[section]

\newtheorem{observation}[theorem]{Observation}

\ifdefined\isarxiv

\else
\renewcommand\cite\citep
\fi

\input{_1_maths}

\begin{document}

\ifdefined\isarxiv

\date{}
\title{\paperTitle}
\author{\paperAuthor}

\else

\title{\paperTitle}

\author{Antiquus S.~Hippocampus, Natalia Cerebro \& Amelie P. Amygdale \thanks{ Use footnote for providing further information
about author (webpage, alternative address)---\emph{not} for acknowledging
funding agencies.  Funding acknowledgements go at the end of the paper.} \\
Department of Computer Science\\
Cranberry-Lemon University\\
Pittsburgh, PA 15213, USA \\
\texttt{\{hippo,brain,jen\}@cs.cranberry-lemon.edu} \\
\And
Ji Q. Ren \& Yevgeny LeNet \\
Department of Computational Neuroscience \\
University of the Witwatersrand \\
Joburg, South Africa \\
\texttt{\{robot,net\}@wits.ac.za} \\
\AND
Coauthor \\
Affiliation \\
Address \\
\texttt{email}
}

%

\newcommand{\fix}{\marginpar{FIX}}
\newcommand{\new}{\marginpar{NEW}}

\maketitle

\fi

\ifdefined\isarxiv
\begin{titlepage}
  \maketitle
  \begin{abstract}
    \input{00_abstract}

  \end{abstract}
  \thispagestyle{empty}
\end{titlepage}

{\hypersetup{linkcolor=black}
\tableofcontents
}
\newpage

\else

\begin{abstract}
\input{00_abstract}
\end{abstract}

\fi


\input{_2_body}



\newpage
\onecolumn
\appendix

\begin{center}
    \textbf{\LARGE Appendix }
\end{center}

\input{_3_app}

\newpage
\clearpage
\ifdefined\isarxiv
\bibliographystyle{alpha}
\bibliography{ref}
\else
\bibliographystyle{iclr2026_conference}
\bibliography{ref}
\fi

\end{document}

%% file: _0_vars.tex
\def\isarxiv{1}

\def\paperTitle{Your Vision-Language Model Can’t Even Count to 20: Exposing the Failures of VLMs in Compositional Counting}

\def\paperAuthor{
Xuyang Guo\thanks{\texttt{ gxy1907362699@gmail.com}. Guilin University of Electronic Technology.}
\and
Zekai Huang\thanks{\texttt{ zekai.huang.666@gmail.com}. The Ohio State University.}
\and
Zhenmei Shi\thanks{\texttt{ zhmeishi@cs.wisc.edu}. University of Wisconsin-Madison.}
\and
Zhao Song\thanks{\texttt{ magic.linuxkde@gmail.com}. University of California, Berkeley.}
\and
Jiahao Zhang\thanks{\texttt{ ml.jiahaozhang02@gmail.com}.}
}



%% file: _1_maths.tex
\usepackage{booktabs}
\usepackage{multirow}

\renewcommand{\cite}{\citep}

\usepackage[most]{tcolorbox} 

\newtcolorbox{promptboxalt}[2][]{ 
    breakable,
    enhanced,
    colback=gray!8,
    colframe=gray!90,
    boxrule=0.8pt,
    arc=2pt,
    boxsep=3pt,
    left=8pt,right=8pt,top=8pt,bottom=8pt,
    fonttitle=\bfseries\normalfont,
    fontupper=\footnotesize,  
    title=#2,
    break at=-\baselineskip,      %
    height fixed for=none,        %
    pad at break*=3mm,           %
    overlay broken={\draw[gray!90,line width=0.8pt] 
        (frame.south west) -- (frame.south east);}, %
    #1
} 

\tcolorboxenvironment{observation}{
  breakable,
  colback=black!10,
  colframe=white,%
  width=\dimexpr\linewidth+10pt\relax,%
  enlarge left by=-5pt,%
  enlarge right by=-5pt,%
  boxsep=5pt,%
  boxrule=0pt,
  left=0pt,right=0pt,top=0pt,bottom=0pt,
  sharp corners,
  before skip=\topsep,
  after skip=\topsep
} 

\newtheorem{question}{Question}
\tcolorboxenvironment{question}{
  breakable,
  colback=black!10,
  colframe=white,%
  width=\dimexpr\linewidth+10pt\relax,%
  enlarge left by=-5pt,%
  enlarge right by=-5pt,%
  boxsep=5pt,%
  boxrule=0pt,
  left=0pt,right=0pt,top=0pt,bottom=0pt,
  sharp corners,
  before skip=\topsep,
  after skip=\topsep
} 

%% file: 00_abstract.tex
Vision-Language Models (VLMs) have become a central focus of today's AI community, owing to their impressive abilities gained from training on large-scale vision-language data from the Web. These models have demonstrated strong performance across diverse tasks, including image understanding, video understanding, complex visual reasoning, and embodied AI. Despite these noteworthy successes, a fundamental question remains: Can VLMs count objects correctly? In this paper, we introduce a simple yet effective benchmark, {\bf VLMCountBench}, designed under a minimalist setting with only basic geometric shapes (e.g., triangles, circles) and their compositions, focusing exclusively on counting tasks without interference from other factors. We adopt strict independent variable control and systematically study the effects of simple properties such as color, size, and prompt refinement in a controlled ablation. 
Our empirical results reveal that while VLMs can count reliably when only one shape type is present, they exhibit substantial failures when multiple shape types are combined (i.e., compositional counting). This highlights a fundamental empirical limitation of current VLMs and motivates important directions for future research.

%% file: _2_body.tex

\input{01_intro}
\input{02_related_works}
\input{03_benchmark}
\input{04_experiments}

\input{05_prompt_refinement}
\input{06_conclusion}

%% file: 01_intro.tex
\section{Introduction}

Vision-Language Models (VLMs) have recently emerged as one of the most influential paradigms in artificial intelligence~\cite{gemma25,wbt+24,openai24}. By jointly training on large-scale paired data from the web, VLMs have demonstrated impressive generalization across a wide range of tasks, including image captioning, video understanding, visual question answering, visual reasoning, and embodied AI~\cite{dxs+23,cyf+24,wzzy24}. These models form the foundation of many recent multimodal systems and are increasingly deployed in real-world applications. Their ability to align vision and language representations in a unified framework has positioned them as a strong foundation for multimodal research and practice.

Despite these remarkable successes, a fundamental question persists: Do VLMs possess reliable basic perceptual abilities? Among these, counting plays a central role, as it underlies numerous higher-level reasoning skills and everyday applications. Counting is both a simple and fundamental visual task that requires identifying discrete objects and enumerating them accurately. Prior work has already raised concerns in related domains. Generative models, for instance, often fail to produce the correct number of objects in synthetic images~\cite{pss+22,cgh+25,hwrl24} and videos~\cite{ghh+25,shl+25}, and CLIP-based models have been shown to struggle with distinguishing and enumerating multiple objects in classification and retrieval settings~\cite{jlc23,pet+23,zlfx24}. However, the specific counting ability of VLMs remains less systematically explored. This motivates our research question: 
\begin{question}
Can state-of-the-art VLMs reliably perform simple and compositional counting tasks?
\end{question}

While some existing benchmarks touch on VLMs' ability to count, they typically do so in complex or noisy environments~\cite{lgg+24,lwh+24,xsz+24}. For example, datasets designed for visual question answering or captioning may contain counting-related queries, but these are embedded within broader tasks involving recognition, commonsense reasoning, or natural image understanding. As a result, it is difficult to disentangle whether a model’s failure arises from counting itself or from unrelated challenges. Similarly, large-scale natural image benchmarks (e.g., COCO~\cite{lmb+14} object detection dataset with labels on the quantity of objects) introduce uncontrolled variability, making it nearly impossible to isolate the exact conditions that cause performance degradation. Thus, despite progress, there remains no controlled and minimalist benchmark dedicated specifically to testing counting in VLMs.

To address this gap, we introduce {\bf VLMCountBench}, a benchmark designed under a strictly minimalist setting. The benchmark consists of simple geometric shapes (e.g., triangles, circles) and their compositions, thereby removing semantic complexity and focusing exclusively on counting. This setting allows us to implement precise variable control, systematically manipulating factors such as color, size, and prompt refinement. By conducting ablation studies under these conditions, we can rigorously analyze VLM performance and identify the specific challenges that lead to counting failures.

We carry out a comprehensive empirical evaluation across multiple state-of-the-art VLMs~\cite{cbs+25,openai24,wbt+24}, covering both open-source and commercial private models, focusing on both single-shape and multi-shape settings. Our results reveal several striking findings:
\begin{itemize}
    \item VLMs can count reliably when only a single shape type is present, achieving high accuracy in simple counting scenarios. 
    \item VLMs exhibit substantial failures in {\bf compositional counting}, where two or more shape types coexist. These failures persist even when the task involves small numbers of objects and minimal visual complexity. 
    \item Performance deteriorates consistently across variations in color, size, and prompt refinement, indicating a lack of stability to simple visual properties. 
\end{itemize}

{\bf Roadmap.} In Section~\ref{sec:related}, we review the related works. In Section~\ref{sec:bench}, we present our proposed benchmark. In Section~\ref{sec:experiments}, we present the main experimental results. We introduce the prompt refinement in Section~\ref{sec:prompt_refinement}. In Section~\ref{sec:conclusion}, we conclude our paper. 

%% file: 02_related_works.tex
\section{Related Works} \label{sec:related}

{\bf Vision-Language Models.}
Motivated by the impressive success of Large language models (LLMs)~\cite{bmr+20, wbz+22, tli+23, chl+24}, scholarly attention is progressively shifting toward the exploration and development of vision-language models, as they have the potential to connect vision and language, achieve more natural human-computer interaction~\cite{kld25}, and advance tasks such as visual question answering~\cite{lcm+23, kjs+25} and multimodal reasoning~\cite{lwlz24,ctg+24}. One significant leap in this area is the revolutionary Visual ChatGPT~\cite{wyq+23}, which combines the reasoning ability of language models with several visual models to achieve natural language-driven image generation, editing, and understanding. Besides, PaLM-E~\cite{dxs+23} has effectively integrated text and vision, achieving remarkable results across a variety of tasks~\cite{xmyr16, mrfm19}. Flamingo~\cite{adl+22} integrates frozen large language models with visual encoders through cross-attention layers, achieving few-shot learning for visual language tasks. Conversely, BLIP2~\cite{llsh23} effectively connects frozen Large Language Models (LLMs) with visual input through a lightweight Q-Former module, which converts image features into a format that LLMs can understand. This design enables high performance in various tasks with minimal additional training. Well-known models such as InstructBLIP~\cite{mrfm19} and LLaVA~\cite{llwl23} have significantly advanced the field by introducing diverse visual instruction-tuning datasets. While prior vision-language models have demonstrated impressive performance across diverse multimodal tasks, their ability to perform precise quantitative analysis on images remains largely unexplored. To address this gap, we propose VLMCountBench to offer insights into their numerical understanding in visual scenes.

{\bf Benchmarks for Vision-Language Models.} 
With the rapid development of Vision-Language Models (VLMs), researchers designed some benchmarks such as TextVQA~\cite{sns+19}, GQA~\cite{hm19}, and DocVQA~\cite{mkj21} to evaluate the ability of VLMs on individual tasks. However, while these task-specific benchmarks provide valuable insights, they do not fully reflect the overall capabilities of VLMs in real-world applications. Therefore, recent efforts~\cite{hlm+24, ynz+24, dhl+24} have shifted toward developing more comprehensive evaluation benchmarks. Meanwhile, VHELM~\cite{ltw+24} comprehensively evaluates the performance of VLMs in multiple dimensions such as perception, reasoning, multilingual ability, and robustness. In addition, several representative benchmarks have been proposed to target different aspects of multimodal evaluation. For example, Perception Test~\cite{pdzc23} focuses on measuring fine-grained perceptual capacity such as color, shape, and size. LVLM eHub~\cite{xsz+24} combines multiple comprehensive benchmarks to design an evaluation platform that covers a wide range of multimodal tasks. LLaVA Bench~\cite{llwl23}, LAMM~\cite{ywc+23}, and Touchstone~\cite{byb+23} leverage GPT-based evaluators to assess model outputs, thereby reducing potential biases introduced by human annotators. Beyond general-purpose benchmarks, some works focus on constructing targeted datasets for more objective and fine-grained evaluation of VLM. MME~\cite{cpy+23} and MMBench~\cite{ldz+24} are designed to strengthen the objective evaluation of VLMs by introducing 2,194 true/false questions and 2,974 multiple-choice questions across diverse ability dimensions. Although existing benchmarks effectively evaluate various VLM capabilities, they primarily target concrete visual entities (e.g., objects, scenes) and largely ignore numerical counting in visual contexts, which motivates the creation of {\bf VLMCountBench}.

{\bf Fundamental Limits of Foundation Models.} 
Studying the fundamental capability limitations of foundation models, for vision-language models and beyond, has long been a central focus of modern AI research, with many theoretical analysis frameworks applied to this problem. Circuit complexity is one of the most prevailing frameworks for bounding the expressive limits of foundation models~\cite{v99,ab09,fzg+23,llzm24}, where a model that can be simulated with a circuit of a lower complexity class (e.g., $\mathsf{TC}^0$) cannot solve a problem that is harder than this class (e.g., $\mathsf{NC}^1$). These results have been used to show that Transformers~\cite{lag+22,cll+25_rope,cll+24_tensor_tc} and their variants~\cite{cll+25_mamba_tc,lll+24_hopfield_tc}, vision models~\cite{kll+25_tc,gkl+25,ccsz25}, and graph learning models~\cite{g24,cgws24,lls+25_graph_tc} exhibit fundamental expressive limitations. 
Another framework is the provably efficient criteria~\cite{as23,as24_iclr,as24_neurips,as25_rank}, which shows that the attention computation in foundation models cannot be approximated with low numerical error under fast computation unless certain conditions hold (e.g., bounded element entries). These results have proved highly useful in analyzing Transformers~\cite{chl+24_rope,lssz24_tat,hwg+24,as25_rope} and their variants~\cite{hlsl24}, Low-Rank Adaptation~\cite{hsk+24}, and diffusion models~\cite{hwsl24,kll+25_fast_var,ccsz25}. 
More fundamental limits have recently emerged, including but not limited to universal approximation~\cite{hwg+24,lhsl25,cll+25_var}, statistical rates~\cite{hwsl24,hwl+24,cmfw24}, lower bounds for optimization~\cite{ks25,cssz25,hzs25+}, and in-context learning~\cite{wsh+24,swxl24,whhz+25,hlzl25}. These theoretical results are also closely connected to empirical findings, such as illusions of reasoning in thinking models~\cite{syz25,ghsz25}, counting limits~\cite{jlc23,bts+24,cgh+25,ghh+25} of diffusion generative models, physical constraints~\cite{lhy+24,ghs+25_physical,cgs+25}, and text manipulation in text-to-video and text-to-image models~\cite{llq+24,pbsj24,ghs+25_text}. 
In this paper, we identify a new fundamental limit of foundation models, with a specific focus on counting in vision-language models.

%% file: 03_benchmark.tex
\section{Benchmark}\label{sec:bench}

In Section~\ref{sec:models}, we introduce the evaluated models in this benchmark. 
In Section~\ref{sec:prompts}, we present the prompts to evaluate the vision language models. In Section~\ref{sec:metrics}, we show the metrics used in this paper.

\subsection{Evalutaed Models}\label{sec:models}

\begin{table}[!ht]
    \centering
    \caption{\textbf{Key Details of the Large Vision-Language
Models.} Gemini-2.5 is a closed-source model that does not provide any information about its parameters.}
    \resizebox{0.8\linewidth}{!}{ 
    \begin{tabular}{|c|c|c|c|c|c|}
        \hline
        \textbf{Model} & \textbf{Source} & \textbf{Year} & \textbf{\# Output Tokens} & \textbf{\# Params} \\
        \hline
        Gemini 2.5 Flash& \cite{cbs+25} & 2025  & 64k & N/A  \\
        \hline
        GPT-4o& \cite{openai24} & 2024  & 16K & 200B  \\
        \hline
        Ernie 4.5& \cite{ernie} & 2025  & 16k & 47B  \\
        \hline
        GLM 4.5v& \cite{glm} & 2025  & 64k & 12B  \\
        \hline
        
        Gemma 3 27B& \cite{gemma25} & 2025  & 128k & 27B  \\
        \hline
        Qwen 2.5 72B& \cite{yyz+25} & 2025  & 32K & 72B  \\
        \hline
        Kimi VL A3B& \cite{kimi} & 2024  & 32K & 3B  \\
        \hline
        Llama 4 Maverick& \cite{llama4} & 2025  & 4K & 17B  \\
        \hline

    \end{tabular}
    }
    \label{tab:models}
\end{table}

We evaluate eight state-of-the-art language models via the OpenRouter API, using their default context lengths and provider settings without any manual adjustment. All inference runs were performed without chain-of-thought prompting; however, Kimi VL A3B~\cite{kimi} and Llama 4 Maverick~\cite{llama4} inherently expose chain-of-thought style reasoning that cannot be disabled, so any intermediate reasoning was ignored and only final outputs were considered. 

\textbf{Open-source models}. Gemma 3 27B~\cite{gemma25} and Qwen 2.5 72B~\cite{yyz+25} provide long-context handling (default capacities of roughly 128 k and 32 k tokens respectively) and support high-resolution images where applicable. Kimi VL A3B~\cite{kimi}, a lightweight 3B parameter vision-language model, and Llama 4 Maverick~\cite{llama4}, a 17B parameter text-focused model with a 4k token window, represent smaller, more agile configurations. Ernie 4.5 47B~\cite{ernie} and GLM 4.5v 12 B~\cite{glm} extend open-source multimodal capabilities with default 16 k and 64 k generation limits, respectively, and adhere to the common image side maximum of 1024 px established by their providers.

\textbf{Closed-source models}. Gemini 2.5 Flash~\cite{cbs+25}, from Google DeepMind, is optimized for fast multimodal inference with a default 64k token limit and image handling up to 1024 px. GPT-4o~\cite{openai24}, OpenAI’s flagship multimodal system with around 200B parameters, operates under a 16k token default and similar image size constraints. 

For all models open and closed, we did not modify decoding hyperparameters or preset any structured outputs beyond provider defaults, ensuring a consistent evaluation setting across architectures and access modalities.

\subsection{Benchmark Prompts and Input Images} \label{sec:prompts}

\begin{wrapfigure}{r}{0.55\textwidth}
    \centering
    \includegraphics[width=0.95\linewidth]{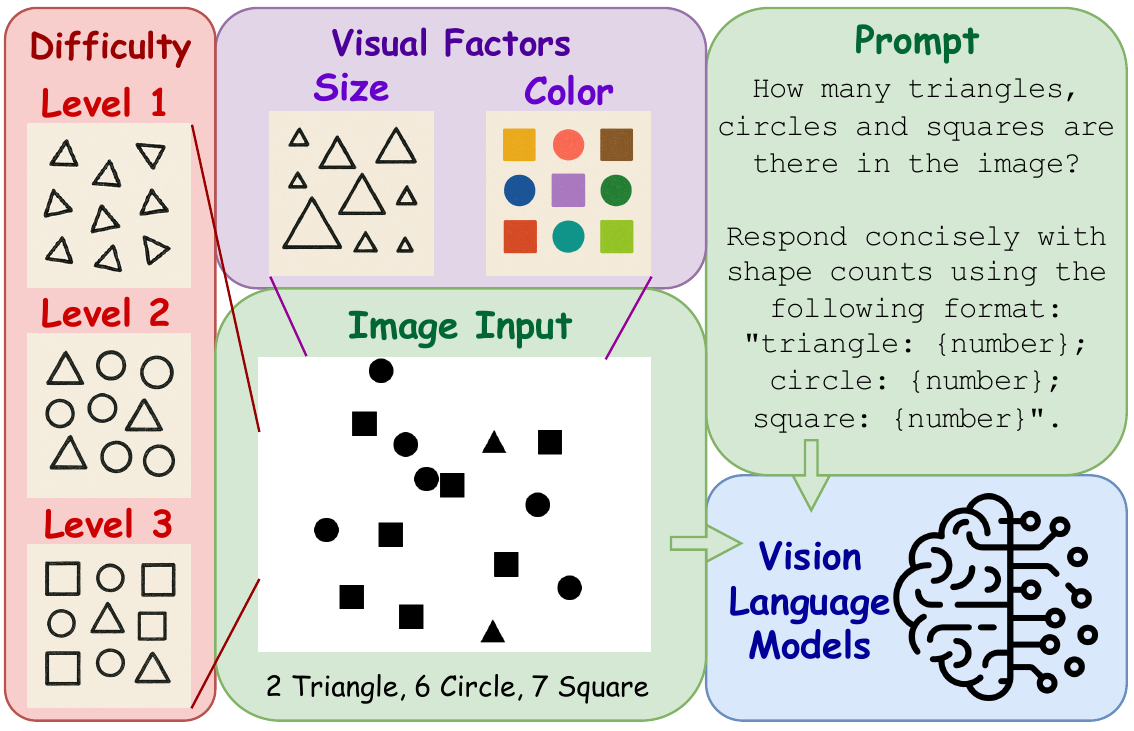}
    \caption{Our experimental design to let VLMs perform object counting.}
    \vskip -0.1in
    \label{fig:vlm_example}
\end{wrapfigure}

Our benchmark is designed to directly evaluate the basic counting ability of vision-language models (VLMs), while minimizing the influence of confounding factors such as complex scene understanding or higher-level reasoning. We adopt a deliberately simple setting where the task is restricted to counting a small number of basic geometric shapes. This allows us to isolate and probe the fundamental ability of VLMs to perform object counting. Despite the simplicity of this setting, we will show that VLMs still exhibit significant failures. The benchmark considers three object types, triangle, square, and circle, and three levels of composition: one, two, or three object types in the same image. For each image, the quantity of objects is sampled between 1 and 20. An illustration of our benchmark is shown in Figure~\ref{fig:vlm_example}..  

{\bf Prompts.} To construct queries, we combine three basic concepts: \textless object\textgreater, \textless level\textgreater~ of composition, and \textless quantity\textgreater. The options are:  
\begin{itemize}
    \item \textless object\textgreater: {\tt `triangle'}, {\tt `square'}, {\tt `circle'}.
    \item \textless level\textgreater: {\tt `level 1'}, {\tt `level 2'}, {\tt `level 3'}.
    \item \textless quantity\textgreater: 1, 2, 3, ..., 20. 
\end{itemize}

In {\tt `level 1'}, the image contains only one type of object. In {\tt `level 2'}, two types of objects are present, and in {\tt `level 3'}, three different types of objects are shown. The corresponding prompt templates are given below:

\begin{promptboxalt}{Level 1 Prompt Template $P_1$}
How many \textless object 1\textgreater~ are there in the image?

Respond concisely with shape counts using the following format: 
``\textless object 1\textgreater: \{number\}''. 
For example: ``\textless object 1\textgreater: 7''. The number 7 is provided as an example only and does not represent the actual quantity of objects in the image.

[image: \textless quantity 1\textgreater~ of \textless object 1\textgreater]
\end{promptboxalt}

\begin{promptboxalt}{Level 2 Prompt Template $P_2$}
How many \textless object 1\textgreater~ and \textless object 2\textgreater~ are there in the image?

Respond concisely with shape counts using the following format: 
``\textless object 1\textgreater: \{number\}; \textless object 2\textgreater: \{number\}''. 
For example: ``\textless object 1\textgreater: 9; \textless object 2\textgreater: 13''. The numbers 7 and 13 are provided as examples only and do not represent the actual quantity of objects in the image.

[image: \textless quantity 1\textgreater~ of \textless object 1\textgreater, \textless quantity 2\textgreater~ of \textless object 2\textgreater]
\end{promptboxalt}

\begin{promptboxalt}{Level 3 Prompt Template $P_3$}
How many \textless object 1\textgreater, \textless object 2\textgreater~ and \textless object 3\textgreater~ are there in the image?

Respond concisely with shape counts using the following format: 
``\textless object 1\textgreater: \{number\}; \textless object 2\textgreater: \{number\}; \textless object 3\textgreater: \{number\}''. For example: ``\textless object 1\textgreater: 3; \textless object 2\textgreater: 11; \textless object 3\textgreater: 6''. The numbers 3, 11, and 6 are provided as examples only and do not represent the actual quantity of objects in the image.

[image: \textless quantity 1\textgreater~ of \textless object 1\textgreater, \textless quantity 2\textgreater~ of \textless object 2\textgreater, \textless quantity 3\textgreater~ of \textless object 3\textgreater]
\end{promptboxalt}

Here, [image: ...] denotes the actual input image containing the specified objects. The placeholders \textless object 1\textgreater, \textless object 2\textgreater~and \textless object 3\textgreater~ always correspond to distinct object types (e.g., a query may ask about triangles and squares, but never triangles and triangles).  

An example prompt at {\tt `level 2'} is shown below:
\begin{promptboxalt}[label=pro:example_1]{Prompt Example 1}
How many {\tt triangles} and {\tt circles} are there in the image?

Respond concisely with shape counts using the following format: 
``{\tt triangles}: \{number\}; {\tt circles}: \{number\}''. 
For example: ``{\tt triangles}: 9; {\tt circles}: 13''. The numbers 9 and 13 are provided as examples only and do not represent the actual quantity of objects in the image.

[image: 7 {\tt triangles}, 15 {\tt circles}]
\end{promptboxalt}

For each level, we randomly sample from all possible combinations of objects and quantities, and retain 200 prompts per level. All images are generated automatically.

{\bf Input Images.} To generate large-scale annotated data, we employ a simple automatic image generator. This can be implemented with basic Python commands, without relying on costly or time-consuming modern generative models, while still being sufficient to reveal the counting limitations of VLMs. Each image is a $640 \times 480$ canvas with a white background and stored as a JPG file. All shapes are drawn with black borders, white interiors, identical sizes, and no rotation. They are placed uniformly at random on the canvas, with no overlaps, ensuring that object counts remain unambiguous and easily verifiable.  

In the base setting, we restrict our benchmark to varying only quantity and composition. More complex properties that may affect counting performance, such as size, color, and overlapping, are deferred to the ablation study.  

\subsection{Evaluation Metrics} \label{sec:metrics}

For each test sample in our benchmark, we use two evaluation metrics: accuracy and relative error. Accuracy measures whether the VLM’s response is exactly correct, while relative error provides a finer-grained evaluation by quantifying how far the prediction deviates from the ground truth. Let a single test sample be denoted as $q := (p, x, y, m)$, where $p$ is the input prompt, $x$ is the input image, $y \in \mathbb{N}_+^{m}$ is the ground-truth vector of object counts, and $m \in \{1,2,3\}$ is the number of object types. For example, in Prompt Example~1 with two object types (triangle and circle) and counts $7$ and $13$, we have $y = [7, 13]^\top$ and $m=2$.  

{\bf Accuracy.} Accuracy evaluates whether the prediction matches the ground truth for each object type. Let ${\cal Q}$ denote the set of test samples of interest (e.g., all {\sf `Level 2'} samples). The metric is defined as:
\begin{align*}
    \mathrm{Accuracy}({\cal Q}) :=  m^{-1}|{\cal Q}|^{-1} \sum_{(p, x, y, m) \in {\cal Q}} \sum_{i=1}^m \boldsymbol{1}[\mathrm{VLM}(p,x)_i = y_i],
\end{align*}
where $\boldsymbol{1}[\cdot]$ is the indicator function, which returns $1$ if the condition inside is true and $0$ otherwise, and $\mathrm{VLM}(p,x) \in \mathbb{N}_+^{m}$ is the predicted object counts.  

Intuitively, for each sample $q$, we compute the fraction of object types predicted exactly correctly, then average over all samples in ${\cal Q}$. For instance, if an image contains three object types (triangle, circle, square) and the model predicts only the square count correctly, then the accuracy for this sample is $1/3$. The final accuracy is the mean of such values over all test samples. 

{\bf Relative Error.} While accuracy captures exact correctness, it does not reflect how close the prediction is when incorrect. To address this, we use relative error, which measures the normalized deviation of predicted counts from ground truth. Formally:
\begin{align*}
    \mathrm{RelativeError}({\cal Q}) :=  m^{-1}|{\cal Q}|^{-1} \sum_{(p, x, y, m) \in {\cal Q}} \sum_{i=1}^m y_i^{-1} \cdot |\mathrm{VLM}(p,x)_i - y_i|,
\end{align*}
where $\mathrm{VLM}(p,x) \in \mathbb{N}_+^{m}$ again denotes the predicted counts.  

This metric computes, for each sample $q$, the average relative error across object types, and then averages over all samples in ${\cal Q}$. For example, if an image contains 16 circles and 10 squares, and the model predicts 8 circles and 8 squares, then the relative error is: $0.5 \cdot (|8-16| / 16+ |8-10| / 10) = 0.5 \cdot (0.5 + 0.2) = 0.35$.  
Thus, relative error provides a more detailed measure of how far predictions deviate from the true counts.  

%% file: 04_experiments.tex
\section{Experiments}\label{sec:experiments}

We present the main experimental results of the VLMCountBench in this section. Specifically, in Section~\ref{sec:comp_counting}, we show the main results on compositional counting. In Section~\ref{sec:perturb}, we present the impact of visual perturbations.

\subsection{Compositional Counting} \label{sec:comp_counting}

\begin{table*}[!ht]
\centering
\caption{{\bf Overall Counting Accuracy and Relative Error Across various Object Types.} The models are listed in a sequence of descending overall count accuracy. We highlight the top 3 models with the best counting accuracy in {\color{blue} blue}, and top 3 models with the least relative error in {\color{red} red}.
}
\resizebox{1\linewidth}{!}{
\begin{tabular}{|c|cccccccc|}
\toprule
\multirow{2}{*}{\centering\textbf{Model}} &  
\multicolumn{2}{c}{\textbf{Level 1}} &      
\multicolumn{2}{c}{\textbf{Level 2}} & 
\multicolumn{2}{c}{\textbf{Level 3}} & 
\multicolumn{2}{c|}{\textbf{Overall}} \\ 
\cmidrule(lr){2-3} \cmidrule(lr){4-5} \cmidrule(lr){6-7} \cmidrule(lr){8-9}
& \textbf{Count Acc} & \textbf{Relative Error} & \textbf{Count Acc} & \textbf{Relative Error} & \textbf{Count Acc} & \textbf{Relative Error} & \textbf{Count Acc} & \textbf{Relative Error} \\
\midrule
Gemma3 27B & 0.26 & 0.14 & 0.21 & 0.23 & 0.22 & 0.25 & 0.23 & 0.21 \\
Kimi VL A3B & 0.29 & 0.23 & 0.22 & 0.27 & 0.19 & 0.30 & 0.23 & 0.27 \\
Llama4 Maverick & 0.38 & 0.15 & 0.33 & 0.14 & 0.25 & 0.19 & 0.32 & 0.16 \\
Gpt-4o & 0.44 & 0.07 & 0.39 & 0.10 & 0.23 & 0.17 & 0.35 & 0.11 \\
Ernie 4.5 & 0.52 & 0.05 & 0.43 & 0.08 & {\color{blue}0.38} & {\color{red}0.10} & 0.44 & 0.08 \\
Gemini 2.5 Flash & {\color{blue}0.58} & {\color{red}0.04} & {\color{blue}0.54} & {\color{red}0.05} & 0.30 & 0.13 & {\color{blue}0.47} &{\color{red}0.07} \\
GLM4.5v & {\color{blue}0.56} & {\color{red}0.05} & {\color{blue}0.49} & {\color{red}0.07} & {\color{blue}0.43} & {\color{red}0.08} & {\color{blue}0.49} & {\color{red}0.07} \\
Qwen2.5 72B& {\color{blue}0.60} & {\color{red}0.04} & {\color{blue}0.56} & {\color{red}0.05} & {\color{blue}0.45} & {\color{red}0.07} & {\color{blue}0.53} & {\color{red}0.05} \\

\bottomrule
\end{tabular}
}
\label{tab:count_main_acc_fid}
\end{table*}

We conduct experiments across three levels: contexts containing one object, two objects, and three objects. For each level, the number of shapes ranges from 1 to 20. Table~\ref{tab:count_main_acc_fid} presents vision-language models' counting performance when varying both the number of object types (one, two, or three) and the number of object instances (ranging from 1 to 20) within the input context.

As shown in Table~\ref{tab:count_main_acc_fid}, current vision-language models still face significant challenges in counting, especially when dealing with multiple objects or diverse object types within the input images. Notably, even the best-performing vision-language model in our benchmark achieves only modest accuracy. For instance, Qwen2.5 72B~\cite{yyz+25} achieved an accuracy of 0.60 at Level 1, but its accuracy substantially declined to 0.45 at Level 3, highlighting the difficulty of the counting task. These findings point to the following insight:

\begin{observation}
    Our results reveal that current vision-language models do not perform ideally on the counting task, and there remains a substantial gap between existing vision-language models' capabilities and the reliable counting ability required for practical applications.
\end{observation}

Across all vision-language models in our benchmark, there is a refined relationship between accuracy and relative error, with relative error serving as a fine-grained metric specifically designed to evaluate counting performance. Even when a model’s prediction is incorrect, a smaller relative error indicates that the predicted counts are closer to the ground truth. In addition, we observed that higher accuracy typically corresponds to smaller relative errors, indicating that models with higher accuracy tend to produce more reliable counting results. For example, Qwen2.5 72B~\cite{yyz+25} has the highest overall counting accuracy at 0.53 and the lowest overall relative error at 0.05. At Level 1, its accuracy is 0.60 with a relative error of 0.04, while at Level 3, the accuracy drops to 0.45 with a slight increase in relative error to 0.07, its relatively small relative error indicates that its counting results are usually close to ground truth, compared to models with lower accuracy and larger relative errors, such as Kimi VL A3B~\cite{kimi}, which has an overall accuracy of 0.23 and a relative error of 0.27, demonstrating a certain degree of counting ability. This brings us a novel insight:

\begin{observation}
    Vision-language models that achieve higher accuracy tend to have smaller relative errors, indicating a stronger counting ability. Conversely, vision-language models with lower accuracy generally show larger relative errors, suggesting limited counting competence. This demonstrates that some vision-language models possess a certain degree of visual counting capability, while others struggle to reliably quantify objects.
\end{observation}

When the number of object types in the input image increases, we observe a clear trend: higher composition levels lead to reduced counting accuracy and increased relative error. For example, Gemini 2.5 Flash achieves a counting accuracy of 0.58 at Level 1, which decreases to 0.54 at Level 2 and further drops to 0.30 at Level 3. Its relative error correspondingly rises from 0.04 to 0.05and then to 0.13. Similar phenomena are observed in GLM4.5v and Qwen2.5 72B, where accuracy declines and relative error rises as more object types are present. From this, we derive the following insight:

\begin{observation}
     Even one of the best-performing models experiences substantial performance degradation as the scene composition becomes more complex. This indicates that current vision-language models may struggle to distinguish multiple object types in a single visual scene, and the interaction between object types (e.g., similar appearances) may further confuse the vision-language models.
\end{observation}

\subsection{Impact of Visual Perturbations} \label{sec:perturb}

\begin{figure*}[!ht]
    \centering
    \includegraphics[width=1\linewidth]{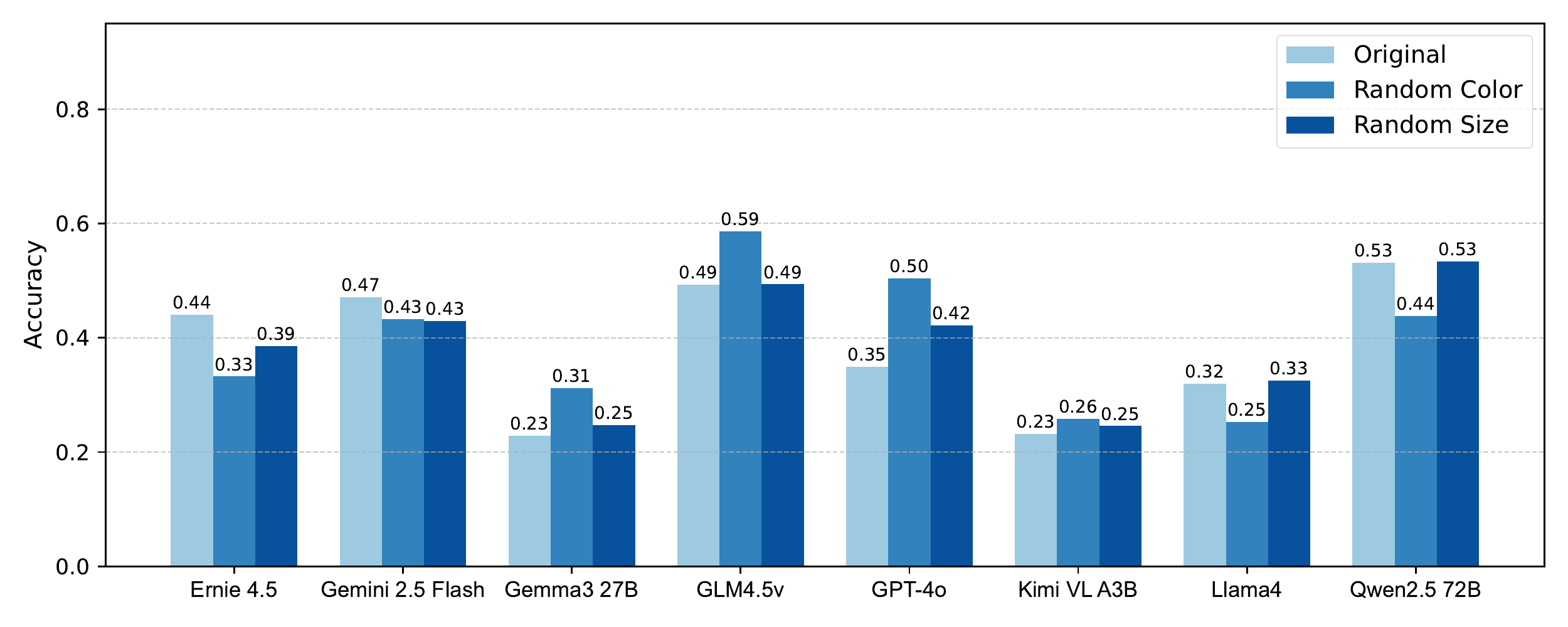}
    \caption{
    {\bf Impact of Visual Perturbations on Model Accuracy}.}
    \label{fig:ablation_accuracy}
\end{figure*}

To better explore the current vision-language models' performance in the counting task. We conduct an ablation study based on our benchmark, VLMCountBench. Figures~\ref{fig:ablation_accuracy} report counting performance, measured by accuracy and relative error, under the three ablation settings. In the original setting, which serves as our main experiment, all shapes are uniform in size and colored black. In the random color setting, shapes are randomly assigned different colors while all other conditions remain identical to the main experiment. In the random size setting, the shape will randomly resize, possibly larger or smaller, with all other conditions remaining unchanged. This setting enables us to systematically evaluate the impact of visual perturbations, such as color and size variations, on the counting performance of vision-language models.

As illustrated in Figure~\ref{fig:ablation_accuracy}, applying random color and random size perturbations to input images leads to varying impacts on counting performance across vision-language models. In particular, GLM4.5v~\cite{glm} and GPT-4o~\cite{openai24} actually benefit from color variations, showing notable increases in accuracy compared with the original setting, possibly because the color differences make objects easier to distinguish. while Ernie 4.5~\cite{ernie} and Qwen2.5 72B~\cite{yyz+25} experience substantial drops, suggesting that these models may rely on specific color distributions learned during training, and that color randomization can disrupt their counting mechanism. In contrast, size perturbations generally cause smaller impacts on performance. Qwen2.5 72B~\cite{yyz+25} and GLM4.5v~\cite{glm} remain relatively high accuracy, while Gemma3 27B~\cite{gemma25} and Kimi VL A3B~\cite{kimi} continue to perform at lower levels. Based on the above analysis, we make the following observations:

\begin{observation}
   Perturbations in color and size could positively or negatively affect counting performance, and the majority of vision-language models are more sensitive to color changes than to size variations, reflecting the different robustness features between vision-language models.
\end{observation}

%% file: 05_prompt_refinement.tex
\section{Prompt Refinement}\label{sec:prompt_refinement}

In this subsection, we evaluate whether the counting limitations of VLMs can be simply resolved by prompt refinements. In Section~\ref{sec:refinement_prompt}, we illustrate the prompt refinement in our work. We present the prompt refinement result and discuss the current discoveries regarding the counting capability of VLMs in Section~\ref{sec:result_discuss}.

\subsection{The Proposed Prompts}\label{sec:refinement_prompt}

Let the prompt template for the three difficulty levels in Section~\ref{sec:prompts} be $P_1, P_2, P_3$. In this section, we introduce several refinement prompts that hint the VLMs to solve the complex counting task by task decomposition, splitting the original task into smaller and manageable parts. These refinement prompts are denoted as $P_{r,1}$ and $P_{r,2}$, and our final prompt used to evaluate the VLMs is denoted by $P ~ || ~ P_r$, where $||$ represents concatenation.

Specifically, $P_r$ has several instantiations. 

{\bf Spatial Decomposition.} 
We found that directly requiring the VLMs to provide a global number may result in omissions or duplications in image counting tasks. Inspired by this, we designed a spatial decomposition approach that breaks down counting tasks into spatial dimensions. We demand VLMs first count the number of objects in the left half of the image, then count the right half, and finally add the results of the two parts. We believe that such prompt refinement can help the VLMs form a local-global inference process, thereby improving the counting performance. Our prompt can be shown as follows:

\begin{promptboxalt}{Spatial Decomposition Prompt $P_{r,1}$}
    First count the objects on the left half of the image, then the right half, and add them together.
\end{promptboxalt}

In specific applications, such as counting triangles and circles in an image, we require the VLMs to "count the left first, then the right, and finally merge the results", and output the quantities of each category in a fixed format. The details example can be shown as follows:

\begin{promptboxalt}{A Level 2 Spatial Decomposition Example $P_2~||~P_{r,1}$}
    How many {\tt triangles} and {\tt circles} are there in the image?

Respond concisely with shape counts using the following format: 
``{\tt triangles}: \{number\}; {\tt circles}: \{number\}''. 
For example: ``{\tt triangles}: 9; {\tt circles}: 13''. The numbers 9 and 13 are provided as examples only and do not represent the actual quantity of objects in the image.

{\color{blue} First count the objects on the left half of the image, then the right half, and add them together.}

[image: 7 {\tt triangles}, 15 {\tt circles}]
\end{promptboxalt}

{\bf Type Decomposition.} 
Another human-inspired method for counting a great number of objects in an image is to first count one category of objects and then proceed to the next. The type decomposition strategy of counting by category could avoid confusion between different categories and improve the counting performance of the VLMs. We define our prompt as follows:

\begin{promptboxalt}{Type Decomposition Prompt $P_{r,2}$}
Count all instances of \textless object 1\textgreater first, then all instances of \textless object 2\textgreater, and then all instances of \textless object 3\textgreater.
\end{promptboxalt}

For example, when the image contains triangles, circles, and squares, we explicitly require the VLMs to "count triangles first, then circles, and finally squares", and provide the results in a unified format. The details example can be shown as follows:

\begin{promptboxalt}{A Level 3 Spatial Decomposition Example $P_3~||~P_{r,2}$}
How many {\tt triangles}, {\tt circles}, and {\tt squares} are there in the image?

Respond concisely with shape counts using the following format: 
``{\tt triangles}: \{number\}; {\tt circles}: \{number\}; {\tt squares}: \{number\}''. 
For example: ``{\tt triangles}: 9; {\tt circles}: 13; {\tt squares}: 6''. The numbers 9, 13, and 6 are provided as examples only and do not represent the actual quantity of objects in the image.

{\color{blue}Count all instances of {\tt triangles} first, then all instances of {\tt circles}, and then all instances of {\tt squares}.}

[image: 7 {\tt triangles}, 15 {\tt circles}, 10 {\tt squares}]
\end{promptboxalt}

\subsection{Results and Discussion}\label{sec:result_discuss}

\begin{table*}[!ht]
\centering
\caption{{\bf Counting Accuracy and Relative Error for Spatial and Type Decomposition.} The models are listed in a sequence of descending overall count accuracy. We highlight the top 3 models with the best counting accuracy in {\color{blue} blue}, and top 3 models with the least relative error in {\color{red} red}.}
\resizebox{1\linewidth}{!}{
\begin{tabular}{|c|cccccc|}
\toprule
\multirow{2}{*}{\centering\textbf{Model}} &  
\multicolumn{2}{c}{\textbf{Original}} &      
\multicolumn{2}{c}{\textbf{Spatial}} & 
\multicolumn{2}{c|}{\textbf{Type}}  \\ 
\cmidrule(lr){2-3} \cmidrule(lr){4-5} \cmidrule(lr){6-7} 
& \textbf{Count Acc} & \textbf{Relative Error} & \textbf{Count Acc} & \textbf{Relative Error} & \textbf{Count Acc} & \textbf{Relative Error}  \\
\midrule
Gemma3 27B & 0.26 & 0.14 &  0.30 & 0.15 & 0.16 & 0.49\\
Kimi VL A3B & 0.29 & 0.23 &  0.18 & 0.37 & 0.15 & 0.50\\
Llama4 Maverick & 0.38 & 0.15 & 0.35 & 0.14 & 0.21 & 0.44\\
Gpt-4o & 0.44 & 0.07  & 0.43 & 0.08 & 0.26 & 0.40\\
Ernie 4.5 & 0.52 & 0.05  & 0.43& 0.09 & 0.26 & 0.41\\
Gemini 2.5 Flash & {\color{blue}0.58} & {\color{red}0.04}  & {\color{blue}0.46} & {\color{red}0.07} & {\color{red}0.29} & {\color{blue}0.39}\\
GLM4.5v & {\color{blue}0.56} & {\color{red}0.05}  &{\color{blue}0.46} & {\color{red}0.08} & {\color{red}0.31} & {\color{blue}0.39}\\
Qwen2.5 72B& {\color{blue}0.60} & {\color{red}0.04}  & {\color{blue}0.47}& {\color{red}0.07} & {\color{red}0.35} & {\color{blue}0.38}\\
\bottomrule
\end{tabular}
}
\label{tab:count_main_acc_fid_2}
\end{table*}

Table~\ref{tab:count_main_acc_fid} presents the counting accuracy and relative error under different refinement strategies. The results demonstrate that compared to the original counting prompts, applying spatial decomposition prompts will slightly reduce accuracy and increase relative error. Although the decomposition strategy provides a more structured step-by-step counting process, additional decomposition steps may introduce errors or complicate the inference process, resulting in a slight decrease in counting performance. In contrast, type decomposition exhibits an even larger performance drop in both accuracy and relative error, demonstrating that for current VLMs, dividing by object type will introduce greater noise in the counting process.

%% file: 06_conclusion.tex
\section{Conclusion} \label{sec:conclusion}

In our study, we propose {\bf VLMCountBench}, a novel benchmark specifically designed to evaluate the counting ability of vision-language models under controlled, minimalist settings. Through systematic experiments on a series of state-of-the-art vision-language models, we found that current vision-language models face significant difficulties in accurately calculating objects in input images, especially in compositional counting scenarios involving multiple object types with varying attributes, such as size and color. These results reveal the fundamental limitations of existing vision-language models and emphasize the necessity of future research to enhance robust counting capabilities. We hope that {\bf VLMCountBench} can provide valuable experience for future researchers to develop more accurate and reliable vision-language models.

%% file: _3_app.tex

\input{50_append_impl_details}
\input{51_additional_experiment}

%% file: 50_append_impl_details.tex
{\bf Roadmap.}
Section~\ref{sec:append_impl_detail} shows the model details of ten baseline vision-language models. Section~\ref{sec:additonal_experiments} present additonal experiments.

\section{Model Details}\label{sec:append_impl_detail}

We present further details of vision-language models in this section.

\begin{itemize}
    \item {\bf GPT 4o}~\cite{openai24}: Created by the OpenAI in 2024, GPT-4o is a closed-source multimodal model. GPT 4o integrates visual and language processing into a unified architecture, enabling tasks such as image understanding, multimodal reasoning, and interactive dialogue. The model supports multimodal inputs, including text, images, and audio, and it can generate outputs across modalities at a breakneck speed based on the problem. 
    \item {\bf Gemma 3}~\cite{gemma25}: Developed by Google DeepMind and released in 2025. Gemma 3 is an open-source vision-language model. It supports multimodal inputs, allowing users to combine text and images within a single prompt. It supports over 140 languages and includes built-in safety tools for filtering sensitive visual content.
    \item {\bf Qwen2 VL 72B}~\cite{wbt+24}: Qwen VL 72B is an open-source vision-language model by Alibaba in 2024. It supports multimodal input, including text and images, capable of processing high-resolution images and performing fine-grained understanding.
    \item {\bf Gemini 2.5 Flash}~\cite{cbs+25}: Developed by Google DeepMind in 2025, Gemini 2.5 Flash is a closed-source multimodal model that supports processing text, image, video, and audio inputs. Besides, the model has built-in thinking capabilities to observe its reasoning process during the generation process
    \item {\bf ERNIE 4.5 VL}~\cite{ernie}: ERNIE 4.5 VL is an open-source vision-language model from Baidu in 2025. It can integrate and text and images, providing different modes of thinking and non-thinking, and support long contextual lengths
    \item {\bf GLM 4.5V}~\cite{glm}: GLM 4.5V is an open-source vision-language model released by Zhipu AI in 2025. It is capable of processing multiple types of inputs, including text, images, and video, and it can handle long-context tasks up to 66K tokens with high efficiency and accuracy.
    \item {\bf Kimi VL A3B}~\cite{kimi}: Kimi VL A3B is an open-source vision-language model released by Moonshot AI in 2025. It supports a wide range of multimodal inputs, including text, high-resolution images, short video clips, and optional OCR or GUI inputs. In addition, it supports advanced reasoning using a "thinking mode", including text-guided image editing and style conversion. 
    \item {\bf Llama 4 maverick}~\cite{llama4}: Llama-4-maverick is an open-source vision-language model from Meta. It adopts a Mixture-of-Experts (MoE) architecture with 17B active parameters, enabling efficient support of multimodal input, including text and high-resolution images, and provides a 128K token context window.
\end{itemize}

We also present the pricing details of all the mdoels in Figure~\ref{tab:model_pricing}.

\begin{table}[!ht]
    \centering
    \caption{\textbf{Key Details of the Large Vision-Language
Models.} (Free models up to 1000 requests per day)}
    \resizebox{0.95\linewidth}{!}{ 
    \begin{tabular}{|c|c|c|c|c|c|}
        \hline
        \textbf{Model} & \textbf{free access?} & \textbf{price/prompt} & \textbf{Token Price}  \\
        \hline
        
        Gemini 2.5 Flash& No & \$0.004 & \$0.30/M input  \$2.50/M output  \$1.238/K input imgs   \\
        \hline
        GPT-4o & No & \$0.005 & \$5/M input \$15/M output \$7.225/K input imgs \\
        \hline
        ERNIE 4.5 & No & \$0.0007 & \$0.14/M input \$0.56/M output  \\
        \hline
        GLM 4.5V & No & \$0.001 & \$0.5/M input \$1.8/M output  \\
        \hline
        Gemma 3 27B& Yes &\$0.00005  & \$0.067/M input \$0.267/M output  \\
        \hline
        Qwen 2.5 72B & Yes &  \$0.0001 &  \$0.25/M input \$0.75/M output  \\
        \hline
        Kimi VL A3B & Yes & \$0.0001& \$0.025/M input \$0.1/M output \\
        \hline
        Llama 4 Maverick & Yes & \$0.0003&  \$0.15/M input \$0.6/M output  \$0.668/K input imgs \\
        \hline

    \end{tabular}
    }
    \label{tab:model_pricing}
\end{table}

%% file: 51_additional_experiment.tex
\section{Additional Experiments}\label{sec:additonal_experiments}

\begin{figure*}[!ht]
    \centering
    \includegraphics[width=1\linewidth]{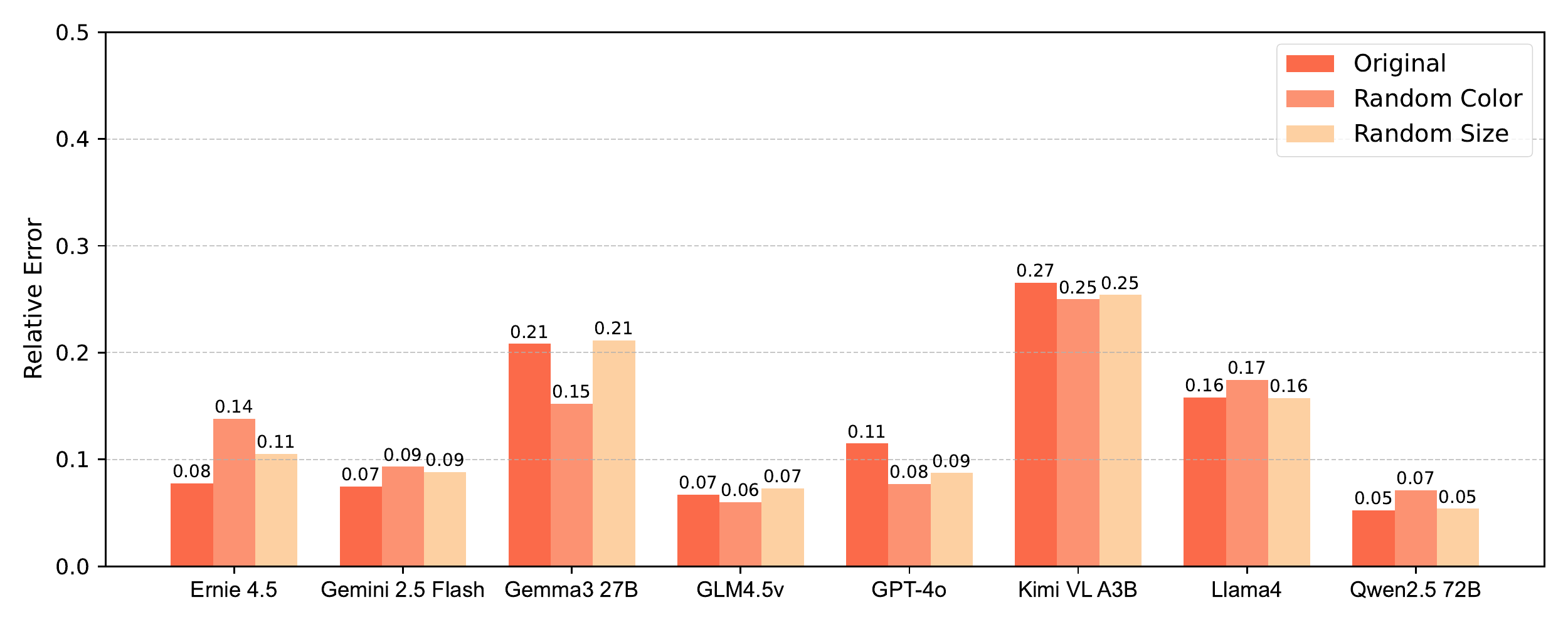}
    \caption{
    {\bf Impact of Visual Perturbations on Model Relative Error}.}
    \label{fig:ablation_error}
\end{figure*}

Due to space constraints, Figure~\ref{fig:ablation_error} has been moved here.

%% file: main.bbl
\newcommand{\etalchar}[1]{$^{#1}$}
\begin{thebibliography}{WZZYL24}

\bibitem[AB09]{ab09}
Sanjeev Arora and Boaz Barak.
\newblock {\em Computational complexity: a modern approach}.
\newblock Cambridge University Press, 2009.

\bibitem[ADL{\etalchar{+}}22]{adl+22}
Jean-Baptiste Alayrac, Jeff Donahue, Pauline Luc, Antoine Miech, Iain Barr, Yana Hasson, Karel Lenc, Arthur Mensch, Katherine Millican, Malcolm Reynolds, et~al.
\newblock Flamingo: a visual language model for few-shot learning.
\newblock In {\em Advances in Neural Information Processing Systems}, 2022.

\bibitem[AS23]{as23}
Josh Alman and Zhao Song.
\newblock Fast attention requires bounded entries.
\newblock In {\em Advances in Neural Information Processing Systems(NeurIPS)}, 2023.

\bibitem[AS24a]{as24_neurips}
Josh Alman and Zhao Song.
\newblock The fine-grained complexity of gradient computation for training large language models.
\newblock In {\em NeurIPS}, 2024.

\bibitem[AS24b]{as24_iclr}
Josh Alman and Zhao Song.
\newblock How to capture higher-order correlations? generalizing matrix softmax attention to kronecker computation.
\newblock In {\em ICLR}, 2024.

\bibitem[AS25a]{as25_rope}
Josh Alman and Zhao Song.
\newblock Fast rope attention: Combining the polynomial method and fast fourier transform.
\newblock In {\em arXiv preprint arXiv:2505.11892}, 2025.

\bibitem[AS25b]{as25_rank}
Josh Alman and Zhao Song.
\newblock Only large weights (and not skip connections) can prevent the perils of rank collapse.
\newblock {\em arXiv preprint arXiv:2505.16284}, 2025.

\bibitem[Bai25]{ernie}
ERNIE~Team Baidu.
\newblock Ernie 4.5 technical report, 2025.

\bibitem[BMR{\etalchar{+}}20]{bmr+20}
Tom~B Brown, Benjamin Mann, Nick Ryder, Melanie Subbiah, Jared Kaplan, Prafulla Dhariwal, Arvind Neelakantan, Pranav Shyam, Girish Sastry, Amanda Askell, et~al.
\newblock Language models are few-shot learners.
\newblock In {\em Proceedings of the 34th International Conference on Neural Information Processing Systems}, pages 1877--1901, 2020.

\bibitem[BTS{\etalchar{+}}24]{bts+24}
Lital Binyamin, Yoad Tewel, Hilit Segev, Eran Hirsch, Royi Rassin, and Gal Chechik.
\newblock Make it count: Text-to-image generation with an accurate number of objects.
\newblock {\em arXiv preprint arXiv:2406.10210}, 2024.

\bibitem[BYB{\etalchar{+}}23]{byb+23}
Shuai Bai, Shusheng Yang, Jinze Bai, Peng Wang, Xingxuan Zhang, Junyang Lin, Xinggang Wang, Chang Zhou, and Jingren Zhou.
\newblock Touchstone: Evaluating vision-language models by language models.
\newblock {\em arXiv preprint arXiv:2308.16890}, 2023.

\bibitem[CBS{\etalchar{+}}25]{cbs+25}
Gheorghe Comanici, Eric Bieber, Mike Schaekermann, Ice Pasupat, Noveen Sachdeva, Inderjit Dhillon, Marcel Blistein, Ori Ram, Dan Zhang, Evan Rosen, et~al.
\newblock Gemini 2.5: Pushing the frontier with advanced reasoning, multimodality, long context, and next generation agentic capabilities.
\newblock {\em arXiv preprint arXiv:2507.06261}, 2025.

\bibitem[CCSZ25]{ccsz25}
Yang Cao, Yubin Chen, Zhao Song, and Jiahao Zhang.
\newblock Towards high-order mean flow generative models: Feasibility, expressivity, and provably efficient criteria.
\newblock {\em arXiv preprint arXiv:2508.07102}, 2025.

\bibitem[CGH{\etalchar{+}}25]{cgh+25}
Yuefan Cao, Xuyang Guo, Jiayan Huo, Yingyu Liang, Zhenmei Shi, Zhao Song, Jiahao Zhang, and Zhen Zhuang.
\newblock Text-to-image diffusion models cannot count, and prompt refinement cannot help.
\newblock {\em arXiv preprint arXiv:2503.06884}, 2025.

\bibitem[CGS{\etalchar{+}}26]{cgs+25}
Yubin Chen, Xuyang Guo, Zhenmei Shi, Zhao Song, and Jiahao Zhang.
\newblock T2vworldbench: A benchmark for evaluating world knowledge in text-to-video generation.
\newblock In {\em WACV}, 2026.

\bibitem[CGWS24]{cgws24}
Guanyu Cui, Yuhe Guo, Zhewei Wei, and Hsin-Hao Su.
\newblock Rethinking gnn expressive power from a distributed computational model perspective.
\newblock {\em arXiv preprint arXiv:2410.01308}, 2024.

\bibitem[CHL{\etalchar{+}}24a]{chl+24_rope}
Yifang Chen, Jiayan Huo, Xiaoyu Li, Yingyu Liang, Zhenmei Shi, and Zhao Song.
\newblock Fast gradient computation for rope attention in almost linear time.
\newblock {\em arXiv preprint arXiv:2412.17316}, 2024.

\bibitem[CHL{\etalchar{+}}24b]{chl+24}
Hyung~Won Chung, Le~Hou, Shayne Longpre, Barret Zoph, Yi~Tay, William Fedus, Yunxuan Li, Xuezhi Wang, Mostafa Dehghani, Siddhartha Brahma, et~al.
\newblock Scaling instruction-finetuned language models.
\newblock {\em Journal of Machine Learning Research}, 25(70):1--53, 2024.

\bibitem[CLL{\etalchar{+}}25a]{cll+25_rope}
Bo~Chen, Xiaoyu Li, Yingyu Liang, Jiangxuan Long, Zhenmei Shi, Zhao Song, and Jiahao Zhang.
\newblock Circuit complexity bounds for rope-based transformer architecture.
\newblock In {\em EMNLP}, 2025.

\bibitem[CLL{\etalchar{+}}25b]{cll+25_mamba_tc}
Yifang Chen, Xiaoyu Li, Yingyu Liang, Zhenmei Shi, and Zhao Song.
\newblock The computational limits of state-space models and mamba via the lens of circuit complexity.
\newblock In {\em Conference on Parsimony and Learning}. PMLR, 2025.

\bibitem[CLL{\etalchar{+}}25c]{cll+25_var}
Yifang Chen, Xiaoyu Li, Yingyu Liang, Zhenmei Shi, and Zhao Song.
\newblock Fundamental limits of visual autoregressive transformers: Universal approximation abilities.
\newblock In {\em International Conference on Machine Learning}. PMLR, 2025.

\bibitem[CMFW24]{cmfw24}
Minshuo Chen, Song Mei, Jianqing Fan, and Mengdi Wang.
\newblock An overview of diffusion models: Applications, guided generation, statistical rates and optimization.
\newblock {\em arXiv preprint arXiv:2404.07771}, 2024.

\bibitem[CPY{\etalchar{+}}23]{cpy+23}
Fu~Chaoyou, Chen Peixian, Shen Yunhang, Qin Yulei, Zhang Mengdan, Lin Xu, Yang Jinrui, Zheng Xiawu, Li~Ke, Sun Xing, et~al.
\newblock Mme: A comprehensive evaluation benchmark for multimodal large language models.
\newblock {\em arXiv preprint arXiv:2306.13394}, 2023.

\bibitem[CSSZ25]{cssz25}
Bo~Chen, Zhenmei Shi, Zhao Song, and Jiahao Zhang.
\newblock Provable failure of language models in learning majority boolean logic via gradient descent.
\newblock {\em arXiv preprint arXiv:2504.04702}, 2025.

\bibitem[CTG{\etalchar{+}}24]{ctg+24}
Yew~Ken Chia, Vernon Toh, Deepanway Ghosal, Lidong Bing, and Soujanya Poria.
\newblock Puzzlevqa: Diagnosing multimodal reasoning challenges of language models with abstract visual patterns.
\newblock In {\em Findings of the Association for Computational Linguistics: ACL 2024}, pages 16259--16273, 2024.

\bibitem[CYF{\etalchar{+}}24]{cyf+24}
An-Chieh Cheng, Hongxu Yin, Yang Fu, Qiushan Guo, Ruihan Yang, Jan Kautz, Xiaolong Wang, and Sifei Liu.
\newblock Spatialrgpt: Grounded spatial reasoning in vision-language models.
\newblock {\em Advances in Neural Information Processing Systems}, 37:135062--135093, 2024.

\bibitem[DHL{\etalchar{+}}24]{dhl+24}
Rocktim Das, Simeon Hristov, Haonan Li, Dimitar Dimitrov, Ivan Koychev, and Preslav Nakov.
\newblock Exams-v: A multi-discipline multilingual multimodal exam benchmark for evaluating vision language models.
\newblock In {\em Proceedings of the 62nd Annual Meeting of the Association for Computational Linguistics (Volume 1: Long Papers)}, pages 7768--7791, 2024.

\bibitem[DXS{\etalchar{+}}23]{dxs+23}
Danny Driess, Fei Xia, Mehdi~SM Sajjadi, Corey Lynch, Aakanksha Chowdhery, Brian Ichter, Ayzaan Wahid, Jonathan Tompson, Quan Vuong, Tianhe Yu, et~al.
\newblock Palm-e: An embodied multimodal language model.
\newblock In {\em International Conference on Machine Learning}, pages 8469--8488. PMLR, 2023.

\bibitem[DYX{\etalchar{+}}25]{kimi}
Angang Du, Bohong Yin, Bowei Xing, Bowen Qu, Bowen Wang, Cheng Chen, Chenlin Zhang, Chenzhuang Du, Chu Wei, et~al.
\newblock Kimi-vl technical report.
\newblock {\em arXiv preprint arXiv:2504.07491}, 2025.

\bibitem[FZG{\etalchar{+}}23]{fzg+23}
Guhao Feng, Bohang Zhang, Yuntian Gu, Haotian Ye, Di~He, and Liwei Wang.
\newblock Towards revealing the mystery behind chain of thought: a theoretical perspective.
\newblock {\em Advances in Neural Information Processing Systems}, 36:70757--70798, 2023.

\bibitem[Gem25]{gemma25}
Team Gemma.
\newblock Gemma 3 technical report, 2025.

\bibitem[GHH{\etalchar{+}}25]{ghh+25}
Xuyang Guo, Zekai Huang, Jiayan Huo, Yingyu Liang, Zhenmei Shi, Zhao Song, and Jiahao Zhang.
\newblock Can you count to nine? a human evaluation benchmark for counting limits in modern text-to-video models.
\newblock {\em arXiv preprint arXiv:2504.04051}, 2025.

\bibitem[GHS{\etalchar{+}}25a]{ghs+25_physical}
Xuyang Guo, Jiayan Huo, Zhenmei Shi, Zhao Song, Jiahao Zhang, and Jiale Zhao.
\newblock T2vphysbench: A first-principles benchmark for physical consistency in text-to-video generation.
\newblock {\em arXiv preprint arXiv:2505.00337}, 2025.

\bibitem[GHS{\etalchar{+}}25b]{ghs+25_text}
Xuyang Guo, Jiayan Huo, Zhenmei Shi, Zhao Song, Jiahao Zhang, and Jiale Zhao.
\newblock T2vtextbench: A human evaluation benchmark for textual control in video generation models.
\newblock {\em arXiv preprint arXiv:2505.04946}, 2025.

\bibitem[GHSZ25]{ghsz25}
Xuyang Guo, Zekai Huang, Zhao Song, and Jiahao Zhang.
\newblock Too easily fooled? prompt injection breaks llms on frustratingly simple multiple-choice questions.
\newblock {\em arXiv preprint arXiv:2508.13214}, 2025.

\bibitem[GKL{\etalchar{+}}25]{gkl+25}
Chengyue Gong, Yekun Ke, Xiaoyu Li, Yingyu Liang, Zhizhou Sha, Zhenmei Shi, and Zhao Song.
\newblock On computational limits of flowar models: Expressivity and efficiency.
\newblock {\em arXiv preprint arXiv:2502.16490}, 2025.

\bibitem[Gro24]{g24}
Martin Grohe.
\newblock The descriptive complexity of graph neural networks.
\newblock {\em TheoretiCS}, 3, 2024.

\bibitem[HLM{\etalchar{+}}24]{hlm+24}
Irene Huang, Wei Lin, M~Jehanzeb Mirza, Jacob~A Hansen, Sivan Doveh, Victor~Ion Butoi, Assaf Arbelle, Hilde Kuehne, Trevor Darrell, Chuang Gan, et~al.
\newblock Conme: rethinking evaluation of compositional reasoning for modern vlms.
\newblock In {\em Proceedings of the 38th International Conference on Neural Information Processing Systems}, pages 22927--22946, 2024.

\bibitem[HLSL24]{hlsl24}
Jerry Yao-Chieh Hu, Thomas Lin, Zhao Song, and Han Liu.
\newblock On computational limits of modern hopfield models: A fine-grained complexity analysis.
\newblock In {\em Forty-first International Conference on Machine Learning}, 2024.

\bibitem[HLZL25]{hlzl25}
Jerry Yao-Chieh Hu, Hude Liu, Jennifer~Yuntong Zhang, and Han Liu.
\newblock In-context algorithm emulation in fixed-weight transformers.
\newblock {\em arXiv preprint arXiv:2508.17550}, 2025.

\bibitem[HM19]{hm19}
Drew~A Hudson and Christopher~D Manning.
\newblock Gqa: A new dataset for real-world visual reasoning and compositional question answering.
\newblock In {\em Proceedings of the IEEE/CVF conference on computer vision and pattern recognition}, pages 6700--6709, 2019.

\bibitem[HSK{\etalchar{+}}25]{hsk+24}
Jerry Yao-Chieh Hu, Maojiang Su, En-Jui Kuo, Zhao Song, and Han Liu.
\newblock Computational limits of low-rank adaptation (lora) fine-tuning for transformer models.
\newblock In {\em The Thirteenth International Conference on Learning Representations}, 2025.

\bibitem[HWG{\etalchar{+}}25]{hwg+24}
Jerry Yao-Chieh Hu, Wei-Po Wang, Ammar Gilani, Chenyang Li, Zhao Song, and Han Liu.
\newblock Fundamental limits of prompt tuning transformers: Universality, capacity and efficiency.
\newblock In {\em The Thirteenth International Conference on Learning Representations}, 2025.

\bibitem[HWL{\etalchar{+}}24]{hwsl24}
Jerry Yao-Chieh Hu, Weimin Wu, Zhuoru Li, Sophia Pi, , Zhao Song, and Han Liu.
\newblock On statistical rates and provably efficient criteria of latent diffusion transformers (dits).
\newblock {\em Advances in Neural Information Processing Systems}, 38, 2024.

\bibitem[HWL{\etalchar{+}}25]{hwl+24}
Jerry Yao-Chieh Hu, Weimin Wu, Yi-Chen Lee, Yu-Chao Huang, Minshuo Chen, and Han Liu.
\newblock On statistical rates of conditional diffusion transformers: Approximation, estimation and minimax optimality.
\newblock In {\em The Thirteenth International Conference on Learning Representations}, 2025.

\bibitem[HWRL24]{hwrl24}
Xiaofei Hui, Qian Wu, Hossein Rahmani, and Jun Liu.
\newblock Class-agnostic object counting with text-to-image diffusion model.
\newblock In {\em European Conference on Computer Vision}, pages 1--18. Springer, 2024.

\bibitem[HYG{\etalchar{+}}25]{glm}
Wenyi Hong, Wenmeng Yu, Xiaotao Gu, Guo Wang, Guobing Gan, Haomiao Tang, Jiale Cheng, Ji~Qi, Junhui Ji, Lihang Pan, et~al.
\newblock Glm-4.1 v-thinking: Towards versatile multimodal reasoning with scalable reinforcement learning.
\newblock {\em arXiv e-prints}, pages arXiv--2507, 2025.

\bibitem[HZS{\etalchar{+}}25]{hzs25+}
Jerry Yao-Chieh Hu, Xiwen Zhang, Maojiang Su, Zhao Song, and Han Liu.
\newblock Minimalist softmax attention provably learns constrained boolean functions.
\newblock {\em arXiv preprint arXiv:2505.19531}, 2025.

\bibitem[JLC23]{jlc23}
Ruixiang Jiang, Lingbo Liu, and Changwen Chen.
\newblock Clip-count: Towards text-guided zero-shot object counting.
\newblock In {\em Proceedings of the 31st ACM International Conference on Multimedia}, pages 4535--4545, 2023.

\bibitem[KJS{\etalchar{+}}25]{kjs+25}
Hongyeob Kim, Inyoung Jung, Dayoon Suh, Youjia Zhang, Sangmin Lee, and Sungeun Hong.
\newblock Question-aware gaussian experts for audio-visual question answering.
\newblock In {\em Proceedings of the Computer Vision and Pattern Recognition Conference}, pages 13681--13690, 2025.

\bibitem[KLD25]{kld25}
Yewon Kim, Sung-Ju Lee, and Chris Donahue.
\newblock Amuse: Human-ai collaborative songwriting with multimodal inspirations.
\newblock In {\em Proceedings of the 2025 CHI Conference on Human Factors in Computing Systems}, pages 1--28, 2025.

\bibitem[KLL{\etalchar{+}}25a]{kll+25_fast_var}
Yekun Ke, Xiaoyu Li, Yingyu Liang, Zhizhou Sha, Zhenmei Shi, and Zhao Song.
\newblock On computational limits and provably efficient criteria of visual autoregressive models: A fine-grained complexity analysis.
\newblock {\em arXiv preprint arXiv:2501.04377}, 2025.

\bibitem[KLL{\etalchar{+}}25b]{kll+25_tc}
Yekun Ke, Xiaoyu Li, Yingyu Liang, Zhenmei Shi, and Zhao Song.
\newblock Circuit complexity bounds for visual autoregressive model.
\newblock {\em arXiv preprint arXiv:2501.04299}, 2025.

\bibitem[KS25]{ks25}
Juno Kim and Taiji Suzuki.
\newblock Transformers provably solve parity efficiently with chain of thought.
\newblock In {\em The Thirteenth International Conference on Learning Representations}, 2025.

\bibitem[LAG{\etalchar{+}}23]{lag+22}
Bingbin Liu, Jordan~T Ash, Surbhi Goel, Akshay Krishnamurthy, and Cyril Zhang.
\newblock Transformers learn shortcuts to automata.
\newblock In {\em ICLR}, 2023.

\bibitem[LCM{\etalchar{+}}23]{lcm+23}
Weizhe Lin, Jinghong Chen, Jingbiao Mei, Alexandru Coca, and Bill Byrne.
\newblock Fine-grained late-interaction multi-modal retrieval for retrieval augmented visual question answering.
\newblock In {\em Proceedings of the 37th International Conference on Neural Information Processing Systems}, pages 22820--22840, 2023.

\bibitem[LDZ{\etalchar{+}}24]{ldz+24}
Yuan Liu, Haodong Duan, Yuanhan Zhang, Bo~Li, Songyang Zhang, Wangbo Zhao, Yike Yuan, Jiaqi Wang, Conghui He, Ziwei Liu, et~al.
\newblock Mmbench: Is your multi-modal model an all-around player?
\newblock In {\em European conference on computer vision}, pages 216--233. Springer, 2024.

\bibitem[LGG{\etalchar{+}}24]{lgg+24}
Bohao Li, Yuying Ge, Yixiao Ge, Guangzhi Wang, Rui Wang, Ruimao Zhang, and Ying Shan.
\newblock Seed-bench: Benchmarking multimodal large language models.
\newblock In {\em Proceedings of the IEEE/CVF Conference on Computer Vision and Pattern Recognition}, pages 13299--13308, 2024.

\bibitem[LHSL25]{lhsl25}
Hude Liu, Jerry Yao-Chieh Hu, Zhao Song, and Han Liu.
\newblock Attention mechanism, max-affine partition, and universal approximation.
\newblock {\em arXiv preprint arXiv:2504.19901}, 2025.

\bibitem[LHY{\etalchar{+}}24]{lhy+24}
Jiaxi Lv, Yi~Huang, Mingfu Yan, Jiancheng Huang, Jianzhuang Liu, Yifan Liu, Yafei Wen, Xiaoxin Chen, and Shifeng Chen.
\newblock Gpt4motion: Scripting physical motions in text-to-video generation via blender-oriented gpt planning.
\newblock In {\em Proceedings of the IEEE/CVF conference on computer vision and pattern recognition}, pages 1430--1440, 2024.

\bibitem[LLL{\etalchar{+}}24]{lll+24_hopfield_tc}
Xiaoyu Li, Yuanpeng Li, Yingyu Liang, Zhenmei Shi, and Zhao Song.
\newblock On the expressive power of modern hopfield networks.
\newblock {\em arXiv preprint arXiv:2412.05562}, 2024.

\bibitem[LLQ{\etalchar{+}}24]{llq+24}
Lin Liu, Quande Liu, Shengju Qian, Yuan Zhou, Wengang Zhou, Houqiang Li, Lingxi Xie, and Qi~Tian.
\newblock Text-animator: Controllable visual text video generation.
\newblock {\em arXiv preprint arXiv:2406.17777}, 2024.

\bibitem[LLS{\etalchar{+}}24]{cll+24_tensor_tc}
Xiaoyu Li, Yingyu Liang, Zhenmei Shi, Zhao Song, and Mingda Wan.
\newblock Theoretical constraints on the expressive power of $\mathsf{RoPE}$-based tensor attention transformers.
\newblock {\em arXiv preprint arXiv:2412.18040}, 2024.

\bibitem[LLS{\etalchar{+}}25]{lls+25_graph_tc}
Xiaoyu Li, Yingyu Liang, Zhenmei Shi, Zhao Song, Wei Wang, and Jiahao Zhang.
\newblock On the computational capability of graph neural networks: A circuit complexity bound perspective.
\newblock {\em arXiv preprint arXiv:2501.06444}, 2025.

\bibitem[LLSH23]{llsh23}
Junnan Li, Dongxu Li, Silvio Savarese, and Steven Hoi.
\newblock Blip-2: bootstrapping language-image pre-training with frozen image encoders and large language models.
\newblock In {\em Proceedings of the 40th International Conference on Machine Learning}, pages 19730--19742, 2023.

\bibitem[LLWL23]{llwl23}
Haotian Liu, Chunyuan Li, Qingyang Wu, and Yong~Jae Lee.
\newblock Visual instruction tuning.
\newblock In {\em Proceedings of the 37th International Conference on Neural Information Processing Systems}, pages 34892--34916, 2023.

\bibitem[LLZM24]{llzm24}
Zhiyuan Li, Hong Liu, Denny Zhou, and Tengyu Ma.
\newblock Chain of thought empowers transformers to solve inherently serial problems.
\newblock In {\em The Twelfth International Conference on Learning Representations}, 2024.

\bibitem[LMB{\etalchar{+}}14]{lmb+14}
Tsung-Yi Lin, Michael Maire, Serge Belongie, James Hays, Pietro Perona, Deva Ramanan, Piotr Doll{\'a}r, and C~Lawrence Zitnick.
\newblock Microsoft coco: Common objects in context.
\newblock In {\em European conference on computer vision}, pages 740--755. Springer, 2014.

\bibitem[LSSZ24]{lssz24_tat}
Yingyu Liang, Zhenmei Shi, Zhao Song, and Yufa Zhou.
\newblock Tensor attention training: Provably efficient learning of higher-order transformers.
\newblock {\em arXiv preprint arXiv:2405.16411}, 2024.

\bibitem[LTW{\etalchar{+}}24]{ltw+24}
Tony Lee, Haoqin Tu, Chi~Heem Wong, Wenhao Zheng, Yiyang Zhou, Yifan Mai, Josselin~Somerville Roberts, Michihiro Yasunaga, Huaxiu Yao, Cihang Xie, et~al.
\newblock Vhelm: a holistic evaluation of vision language models.
\newblock In {\em Proceedings of the 38th International Conference on Neural Information Processing Systems}, pages 140632--140666, 2024.

\bibitem[LWH{\etalchar{+}}24]{lwh+24}
Kunchang Li, Yali Wang, Yinan He, Yizhuo Li, Yi~Wang, Yi~Liu, Zun Wang, Jilan Xu, Guo Chen, Ping Luo, et~al.
\newblock Mvbench: A comprehensive multi-modal video understanding benchmark.
\newblock In {\em Proceedings of the IEEE/CVF Conference on Computer Vision and Pattern Recognition}, pages 22195--22206, 2024.

\bibitem[LWLZ24]{lwlz24}
Junlin Lee, Yequan Wang, Jing Li, and Min Zhang.
\newblock Multimodal reasoning with multimodal knowledge graph.
\newblock In {\em Proceedings of the 62nd Annual Meeting of the Association for Computational Linguistics (Volume 1: Long Papers)}, pages 10767--10782, 2024.

\bibitem[Met25]{llama4}
Meta.
\newblock Llama 4, 2025.

\bibitem[MKJ21]{mkj21}
Minesh Mathew, Dimosthenis Karatzas, and CV~Jawahar.
\newblock Docvqa: A dataset for vqa on document images.
\newblock In {\em Proceedings of the IEEE/CVF winter conference on applications of computer vision}, pages 2200--2209, 2021.

\bibitem[MRFM19]{mrfm19}
Kenneth Marino, Mohammad Rastegari, Ali Farhadi, and Roozbeh Mottaghi.
\newblock Ok-vqa: A visual question answering benchmark requiring external knowledge.
\newblock In {\em Proceedings of the IEEE/cvf conference on computer vision and pattern recognition}, pages 3195--3204, 2019.

\bibitem[Ope24]{openai24}
OpenAI.
\newblock Gpt-4o system card, 2024.

\bibitem[PBSJ24]{pbsj24}
Seonmi Park, Inhwan Bae, Seunghyun Shin, and Hae-Gon Jeon.
\newblock Kinetic typography diffusion model.
\newblock In {\em European Conference on Computer Vision}, pages 166--185. Springer, 2024.

\bibitem[PDZC23]{pdzc23}
Viorica Patraucean, Dima Damen, Andrew Zisserman, and Joao Carriera.
\newblock Perception test: A diagnostic benchmark for multimodal video models.
\newblock In {\em Conference on Neural Information Processing Systems}, 2023.

\bibitem[PET{\etalchar{+}}23]{pet+23}
Roni Paiss, Ariel Ephrat, Omer Tov, Shiran Zada, Inbar Mosseri, Michal Irani, and Tali Dekel.
\newblock Teaching clip to count to ten.
\newblock In {\em Proceedings of the IEEE/CVF International Conference on Computer Vision}, pages 3170--3180, 2023.

\bibitem[PSS{\etalchar{+}}22]{pss+22}
Vitali Petsiuk, Alexander~E Siemenn, Saisamrit Surbehera, Zad Chin, Keith Tyser, Gregory Hunter, Arvind Raghavan, Yann Hicke, Bryan~A Plummer, Ori Kerret, et~al.
\newblock Human evaluation of text-to-image models on a multi-task benchmark.
\newblock {\em arXiv preprint arXiv:2211.12112}, 2022.

\bibitem[SHL{\etalchar{+}}25]{shl+25}
Kaiyue Sun, Kaiyi Huang, Xian Liu, Yue Wu, Zihan Xu, Zhenguo Li, and Xihui Liu.
\newblock T2v-compbench: A comprehensive benchmark for compositional text-to-video generation.
\newblock In {\em Proceedings of the Computer Vision and Pattern Recognition Conference}, pages 8406--8416, 2025.

\bibitem[SNS{\etalchar{+}}19]{sns+19}
Amanpreet Singh, Vivek Natarajan, Meet Shah, Yu~Jiang, Xinlei Chen, Dhruv Batra, Devi Parikh, and Marcus Rohrbach.
\newblock Towards vqa models that can read.
\newblock In {\em Proceedings of the IEEE/CVF conference on computer vision and pattern recognition}, pages 8317--8326, 2019.

\bibitem[SWXL24]{swxl24}
Zhenmei Shi, Junyi Wei, Zhuoyan Xu, and Yingyu Liang.
\newblock Why larger language models do in-context learning differently?
\newblock In {\em International Conference on Machine Learning}. PMLR, 2024.

\bibitem[SYZ25]{syz25}
Zhao Song, Song Yue, and Jiahao Zhang.
\newblock Thinking isn't an illusion: Overcoming the limitations of reasoning models via tool augmentations.
\newblock {\em arXiv preprint arXiv:2507.17699}, 2025.

\bibitem[TLI{\etalchar{+}}23]{tli+23}
Hugo Touvron, Thibaut Lavril, Gautier Izacard, Xavier Martinet, Marie-Anne Lachaux, Timoth{\'e}e Lacroix, Baptiste Rozi{\`e}re, Naman Goyal, Eric Hambro, Faisal Azhar, et~al.
\newblock Llama: Open and efficient foundation language models.
\newblock {\em arXiv preprint arXiv:2302.13971}, 2023.

\bibitem[Vol99]{v99}
Heribert Vollmer.
\newblock {\em Introduction to circuit complexity: a uniform approach}.
\newblock Springer Science \& Business Media, 1999.

\bibitem[WBT{\etalchar{+}}24]{wbt+24}
Peng Wang, Shuai Bai, Sinan Tan, Shijie Wang, Zhihao Fan, Jinze Bai, Keqin Chen, Xuejing Liu, Jialin Wang, Wenbin Ge, et~al.
\newblock Qwen2-vl: Enhancing vision-language model's perception of the world at any resolution.
\newblock {\em arXiv preprint arXiv:2409.12191}, 2024.

\bibitem[WBZ{\etalchar{+}}22]{wbz+22}
Jason Wei, Maarten Bosma, Vincent Zhao, Kelvin Guu, Adams~Wei Yu, Brian Lester, Nan Du, Andrew~M Dai, and Quoc~V Le.
\newblock Finetuned language models are zero-shot learners.
\newblock In {\em International Conference on Learning Representations}, 2022.

\bibitem[WHH{\etalchar{+}}25]{whhz+25}
Weimin Wu, Teng{-}Yun Hsiao, Jerry~Yao{-}Chieh Hu, Wenxin Zhang, and Han Liu.
\newblock In-context learning as conditioned associative memory retrieval.
\newblock In {\em Forty-second International Conference on Machine Learning}, 2025.

\bibitem[WSH{\etalchar{+}}25]{wsh+24}
Weimin Wu, Maojiang Su, Jerry Yao-Chieh Hu, Zhao Song, and Han Liu.
\newblock In-context deep learning via transformer models.
\newblock In {\em International Conference on Machine Learning}. PMLR, 2025.

\bibitem[WYQ{\etalchar{+}}23]{wyq+23}
Chenfei Wu, Shengming Yin, Weizhen Qi, Xiaodong Wang, Zecheng Tang, and Nan Duan.
\newblock Visual chatgpt: Talking, drawing and editing with visual foundation models.
\newblock {\em arXiv preprint arXiv:2303.04671}, 2023.

\bibitem[WZZYL24]{wzzy24}
Xiaohan Wang, Yuhui Zhang, Orr Zohar, and Serena Yeung-Levy.
\newblock Videoagent: Long-form video understanding with large language model as agent.
\newblock In {\em European Conference on Computer Vision}, pages 58--76. Springer, 2024.

\bibitem[XMYR16]{xmyr16}
Jun Xu, Tao Mei, Ting Yao, and Yong Rui.
\newblock Msr-vtt: A large video description dataset for bridging video and language.
\newblock In {\em Proceedings of the IEEE Conference on Computer Vision and Pattern Recognition}, pages 5288--5296, 2016.

\bibitem[XSZ{\etalchar{+}}24]{xsz+24}
Peng Xu, Wenqi Shao, Kaipeng Zhang, Peng Gao, Shuo Liu, Meng Lei, Fanqing Meng, Siyuan Huang, Yu~Qiao, and Ping Luo.
\newblock Lvlm-ehub: A comprehensive evaluation benchmark for large vision-language models.
\newblock {\em IEEE Transactions on Pattern Analysis and Machine Intelligence}, 2024.

\bibitem[YNZ{\etalchar{+}}24]{ynz+24}
Xiang Yue, Yuansheng Ni, Kai Zhang, Tianyu Zheng, Ruoqi Liu, Ge~Zhang, Samuel Stevens, Dongfu Jiang, Weiming Ren, Yuxuan Sun, et~al.
\newblock Mmmu: A massive multi-discipline multimodal understanding and reasoning benchmark for expert agi.
\newblock In {\em Proceedings of the IEEE/CVF Conference on Computer Vision and Pattern Recognition}, pages 9556--9567, 2024.

\bibitem[YWC{\etalchar{+}}23]{ywc+23}
Zhenfei Yin, Jiong Wang, Jianjian Cao, Zhelun Shi, Dingning Liu, Mukai Li, Xiaoshui Huang, Zhiyong Wang, Lu~Sheng, Lei Bai, et~al.
\newblock Lamm: language-assisted multi-modal instruction-tuning dataset, framework, and benchmark.
\newblock In {\em Proceedings of the 37th International Conference on Neural Information Processing Systems}, pages 26650--26685, 2023.

\bibitem[YYZ{\etalchar{+}}25]{yyz+25}
An~Yang, Baosong Yang, Beichen Zhang, Binyuan Hui, Bo~Zheng, Bowen Yu, Chengyuan Li, Dayiheng Liu, Fei Huang, Haoran Wei, Huan Lin, Jian Yang, Jianhong Tu, Jianwei Zhang, Jianxin Yang, Jiaxi Yang, Jingren Zhou, Junyang Lin, Kai Dang, Keming Lu, Keqin Bao, Kexin Yang, Le~Yu, Mei Li, Mingfeng Xue, Pei Zhang, Qin Zhu, Rui Men, Runji Lin, Tianhao Li, Tianyi Tang, Tingyu Xia, Xingzhang Ren, Xuancheng Ren, Yang Fan, Yang Su, Yichang Zhang, Yu~Wan, Yuqiong Liu, Zeyu Cui, Zhenru Zhang, and Zihan Qiu.
\newblock Qwen2.5 technical report, 2025.

\bibitem[ZLFX24]{zlfx24}
Zeliang Zhang, Zhuo Liu, Mingqian Feng, and Chenliang Xu.
\newblock Can clip count stars? an empirical study on quantity bias in clip.
\newblock {\em arXiv preprint arXiv:2409.15035}, 2024.

\end{thebibliography}
